\documentclass{article}

\usepackage{PRIMEarxiv}

\usepackage[utf8]{inputenc} % allow utf-8 input
\usepackage[T1]{fontenc}    % use 8-bit T1 fonts
\usepackage{hyperref}       % hyperlinks
\usepackage{url}            % simple URL typesetting
\usepackage{booktabs}       % professional-quality tables
\usepackage{amsfonts}       % blackboard math symbols
\usepackage{nicefrac}       % compact symbols for 1/2, etc.
\usepackage{microtype}      % microtypography
\usepackage{lipsum}
\usepackage{fancyhdr}       % header
\usepackage{graphicx}       % graphics
\graphicspath{{media/}}     % organize your images and other figures under media/ folder
\usepackage{indentfirst}    % indent paragraphs
\usepackage[ruled,linesnumbered]{algorithm2e}  % give those sweet pseudocodes \m/

\title{A Measure for Level of Autonomy Based on Observable System Behavior
}

\author{
  Jason M. Pittman \\
  \texttt{jason.pittman@umgc.edu} \\
}

\begin{document}
\maketitle

\begin{abstract}
Contemporary artificial intelligence systems are pivotal in enhancing human efficiency and safety across various domains. One such domain is autonomous systems, especially in automotive and defense use cases. Artificial intelligence brings learning and enhanced decision-making to autonomy systems’ goal-oriented behaviors and human independence. However, the lack of clear understanding of autonomy system capabilities hampers human-machine or machine-machine interaction and interdiction. This necessitates varying degrees of human involvement for safety, accountability, and explainability purposes. Yet, measuring the level autonomous capability in an autonomous system presents a challenge. Two scales of measurement exist, yet measuring autonomy presupposes a variety of elements not available in the wild. This is why existing measures for level of autonomy are operationalized only during design or test and evaluation phases. No measure for level of autonomy based on observed system behavior exists at this time. To address this, we outline a potential measure for predicting level of autonomy using observable actions. We also present an algorithm incorporating the proposed measure. The measure and algorithm have significance to researchers and practitioners interested in a method to blind compare autonomous systems at runtime. Defense-based implementations are likewise possible because counter-autonomy depends on robust identification of autonomous systems. 
\end{abstract}

% keywords can be removed

\section{Introduction}

Contemporary artificial intelligence (AI) systems play a pivotal role in enhancing human efficiency and safety across various facets of daily life. The overarching objective remains to alleviate human burdens by entrusting artificial systems with tasks deemed either undesirable, dull, or hazardous. Fields such as health, finance, automotive, national defense, and criminal justice have experienced significant impact in this regard.

One area where AI is experiencing rapid innovation is in autonomous systems (Albrecht \& Stone, 2018), particularly in the automotive (Yurtsever et al., 2020) and defense (Reis et al., 2021) industries. While AI is generally understood to be defined by learning and decision-making, autonomous systems (AS) add goal-oriented and independent (from human operators) behaviors (Wang et al, 2019) to the foundational AI definition.  Indeed, at a minimum, AS are capable of self-direction and self-sufficiency (Bradshaw et al., 2013). Extended capabilities are of course desirable and, in a variety of cases, necessary. 

However, according to Bradshaw et al., when then is a lack a clear understanding of highly autonomous system capabilities, the overall system (i.e., human-machine or machine-machine) suffers. More specifically, not recognizing a system’s autonomous capabilities in a timely manner significantly limits the ability to interact with or interdict the system (Scharre \& Horowitz, 2015; Longpre et al., 2022). Whether an AS is driving an automobile or selecting targets for a weapon system, keeping a human in the loop or on the loop is required for safety, accountability, and explainability. In AS development, the degree of human interaction and interdiction possible translates to a level of autonomy (Hudson \& Reeker, 2007; Meakin, 2021). In simple terms, the higher the level of autonomy the less a human is needed.

There is a general problem here because, according to Antsaklis and Rahnama (2018), “[m]easuring the degree of autonomy is non-trivial” (pg 25). In simple terms, designers can specify the level of autonomy for an autonomous system during development, but operators or observers cannot measure the level of autonomy in the wild. Such inability to measure level of autonomy in the wild is a specific problem because emergent runtime behavior by an AS may be less than or greater than the design-stated level of autonomy. Neither case is desirable. Accordingly, the purpose of this study was to develop a measure for predicting the level of autonomy in an observed AS. 

The rest of this work is structured as follows. We provide an overview of existing research related to autonomy, systems with autonomy, and levels of autonomy in the next section. Then, we demonstrate a theoretical framework for measuring level of autonomy in the wild. The framework includes a mathematical expression, an algorithmic abstraction of the expression, and a narrative example thereof. Finally, we discuss the potential significance of the theoretical framework to the field and industry as well as recommendations for future work.

\section{Related Work}

There are four dimensions to the background for this work. The first dimension describes autonomy as a foundational concept and establishes important definitions. The intersection of autonomy and AI (Figure 1) gives rise to autonomous intelligent systems. Here we again establish foundational concepts but also begin to introduce major challenges in the field. For the third, we discuss existing measures for levels of autonomy. Then, lastly, we address explainability and the potential, or lack thereof, for explainability to supplant measures for level of autonomy.

%% insert fig 1
\begin{figure}
    \centering
    \includegraphics{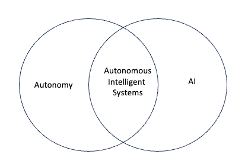}
    \caption{The intersection of autonomy and artificial intelligence.}
    \label{fig:fig1}
\end{figure}

\subsection{Autonomy}
Autonomy is the first dimension related to this study. Autonomy, in this context, provides an operational framework for autonomous intelligent systems and lethal autonomous systems. That is, without autonomy or autonomous systems, neither would exist. This matters because of the growing importance of autonomy in civilian and military contexts.
A persistent issue throughout the literature is defining what is and what is not autonomy (Beer et al., 2014; Williams, 2015; Saxon 2021). Several definitions exist, each providing a material component to our understanding of autonomy. Most commonly we found minor variations on the ability for a system to operate and make decisions (Vladeck 2014; Williams, 2015; Saxon 2021). The variations appeared in how human involvement was articulated in the literature. Thematically, we observed without continuous human involvement, without direct human involvement, and without human input. In all cases, definitions appeared to delimit at what point humans can choose to be on or out of the loop. Thus, human involvement is implicit until it is no longer explicitly required. We conceptualize this definition as the engineering view of autonomy.

We also found the idea that autonomy is the ability of a system to learn and adapt its behavior (Kasabov \& Kozma, 1998; Palanisamy, 2020) which we think of as the computational view of autonomy. Here, the explicit inclusion of learn extends the notion of making decisions. Concurrently, adapt relates to navigating and interacting with an environment. 
In some sense, this definition serves as bridge between fundamental decision-making and the tight coupling between autonomy and the area of operation for such autonomy. On the other side of the definition bridge then, the DoD (Williams, 2015; Saxon, 2021; Ness et al., 2023) positions autonomy as a system capable of executing a pre-planned mission and reacting to uncertainties in the environment without external control. Here, we take the navigation of uncertainty as a differentiating factor from other definitions. Huang et al. (2007) felt similarly when they suggested autonomy “…cannot be separated from its context of mission and environmental complexity” (pg 1). We consider this the defense view of autonomy.

Further issue stems from what constitutes levels of autonomy. On one hand, research (Nickles et al., 2003; Beer et al., 2014; Antsaklis, 2020) suggests autonomy exists on a spectrum. At a minimum, there is a discrete distinction give not all systems possess autonomy and thus are non-autonomous systems. Furthermore, the literature recognizes semi-autonomous systems wherein a human is either in-the-loop or on-the-loop. On the other hand, some researchers such as Bradshaw et al. (2013) argue discrete levels of autonomy are meaningless at best, dangerous to entertain at worst. 

To clarify this work’s position, we find Varela’s (1979) assertion that autonomy is an operationally closed system poignant. When examined in the context of sensing and acting, the autonomy definitions affording for levels of autonomy offered in modern times makes sense. 

Autonomy is operationally closed in scope to the system’s intended function. This does not preclude opportunity for some superordinate functions to be maintained by human operators.  The point is that what functions exist within the system, to the extent the system is autonomous, do operate in closed sensing-acting fashion.

With this in mind, we suggest the field has made progress in terms of understanding what characteristics define an autonomy as well as what still needs to be resolved to achieve full autonomy (i.e., all functions existing within the operationally closed system). Again, if nothing more, the distinction between autonomous and semi-autonomous (citations) establishes levels. In fact, Nickles et al. (2003) asserted knowing the level of autonomy for a given system is a critical need. We found such need to be twofold in the literature. First, as Huang et al. (2007) put forward, levels of autonomy denote system capability in the absence of human direction. Second, Antsaklis (2020) indicated levels of autonomy are vital for comparative purposes when assessing two or more systems. 

Throughout the research, one can find coarse grained scales and models consisting of five (5) levels (SAE, 2016; Antsaklis \& Rahnama, 2018). Fine grained versions exist as well which consist of ten (10) levels. Sheridan, Verplank, and Brooks (1978) offered the foundational taxonomy for this fine-grained model. Later, Endsley and Kaber (1999) revised that scale and situated the 10 levels of autonomy in the perspective of the human operator. Additionally, Hudson and Reeker (2007) described various scales. A final revision came in 2007 as the ALFUS model (Huang et al.). In the majority of, if not all, scales and models, non-autonomous exists as a zero. As well, full autonomy exists as the final level in all scales and models.

It is noteworthy that existing scales for determining or measuring levels of autonomy model systems at design time. We have not found outcomes describing how such modeling may be applicable during runtime (i.e., observationally). Given the scope of such a gap, artificial intelligence represents a potential technological solution.

\subsection{Autonomous intelligent systems}

In simple terms, applying artificial intelligence to autonomous system yields in an autonomous intelligent system (Reis et al., 2021; Chen, Sun, \& Wang, 2022) or AIS. More specifically, autonomous intelligent systems are autonomous systems capable of independently learning and adapting. In that sense, autonomous intelligent systems are the computational view of autonomy manifest. Moreover, alignment in definition of what an autonomous intelligent system is prevents the contention seen elsewhere with describing, comparing, and contrasting types of systems. This appears true across the plethora of AIS research areas such as swarms, robotics, automotive, and defense systems.

The lack of definitional contention may be related to the consensus on what differentiates AIS from mere autonomy. That is, the center of AIS is the supervisory or control mechanism which is where artificial intelligence comes into play. This component replaces the human in or on the loop to a degree commensurate with the level of autonomy. Here, Tsamados and Taddeo (2023) outlined five steps for supervisory control in any autonomous intelligent system: planning, teaching, monitoring, intervening, and learning. Admittedly, the authors indicated one of the two agents involved were human. However, the five steps appear remarkably similar to \textit{self-*} principles at the core of autonomic computing. The connection is more than tenuous. Harel, Marron, and Sifakis (2020) claimed autonomic computing principles serve as the foundation for autonomous intelligent systems.

There are a variety of engaging challenges to solve in AIS. Tyagi and Aswathy (2021) suggested three areas for future research to focus: security, trust, and performance. These problems align with the defense and engineering views of autonomy.  In the computational view, Andresciani and Cingolani (2020) proposed accountability, transparency, and safety. When we delimit challenges related to ethics and responsibility, the literature carries these challenges forward to the very forefront of the field.

\subsection{Measuring autonomy in autonomous systems}

The need to measure the level of autonomy follows logically from the presence of levels of autonomy. Moreover, the need transcends the type and implementation of autonomy. However, as Antsaklis and Rahnama (2018) demonstrated, measuring the level of autonomy in an AS or AIS is not a straightforward exercise. Nevertheless, there exists a variety of quantifiable methods to assess the level of autonomy in a system (Table 1).

%% insert table 1
\begin{table}
    \centering
    \begin{tabular}{ccl}
        \hline
        \textbf{Author(s)} & \textbf{Year} & \textbf{Measure} \\
        \hline
        \hline
         Seth & 2010 & $ga_{x_1 | x_2} = log (var(\xi_{1R(11))} \div ar(\xi_{1U}))$  \\ \\
         Hrabia et al. & 2015 & $PA_{score} , BR_{score} , GP_{score} , L_{score}, C_{norm_i}, = C_{i - min(C_i)}, \div \: max(C_i) $\\ \\ 
         Antsaklis & 2020 & $L = M_g * M_u$ \\ \\ 
         Meakin & 2021 & $BIL \propto  F(N_\Delta state(env) - N_SupervisorInteraction) \div N_\Delta state(env)$ where \\
            & & $BIL \equiv 10^A(N_\Delta state(env) -N_SupervisorInteraction) \div N_\Delta state(env)$
            \\
        \hline
    \end{tabular}
    \caption{Summary of measures for level of autonomy.}
    \label{tab:tab1}
\end{table}

Hrabia et al. (2015) laid groundwork for modern quantification of autonomy by detailing a multi-dimensional autonomy metric framework. The dimensions included adaptation, perception and acting, planning and goals, belief and reasoning, learning, and decision making. Each dimension has its own expression and output metric. Taken together through a relation function, the framework computes a normalized autonomy score. The score is a continuous value which lends a high degree of precision. Yet, Hrabia et al. do not map the individual dimensions or normalized score to scales of autonomy (e.g.,  Huang et al., 2007; Hudson \& Reeker, 2007).

Antsaklis (2020) consolidated the Hrabia et al. (2015) multi-dimensional framework by conceptualizing autonomous systems as control systems. Doing so led to exploring the integration of methods from operations research and AI to achieve higher levels of autonomy. This ultimately led to an expression capable of computing a level of autonomy on the ALFUS (Huang et al., 2007) scale. Notably, Antsaklis derived further expressions incorporating external interventions as well as system performance. 

Meakin (2021) described a method for quantifying the degree of autonomy of a system. The objective was to provide a quantitative basis for comparing two or more systems. Where this method differs, but also connects to level of autonomy scales such as ALFUS (Huang et al., 2007), is in quantifying initiative. Meakin offered imitative as a proxy for independent decision-making. Overall, the measurement is functionally sound because of the reliance on independent behavior. Furthermore, by mapping the measured Behavioural Indepdenence Level (BIL) to the ALFUS Human Independence (HI), Meakin’s method maps to a level of autonomy (Hudson \& Reeker, 2007; Antsaklis, 2020).

Certainly, there have been other proposed quantitative methods related to level of autonomy (Hudson \& Reeker, 2007; Seth, 2010; Nikitenko \& Durst, 2016). Despite the variance in approaches to measuring autonomy, there are several commonalities. Such commonalities include knowledge of designed goals, capabilities, and behaviors. 

Still, none of the existing methods quantify autonomy based on observation in the wild. This appears true despite a variety of measures including some form of behavioral dimension in the autonomy level calculation. Furthermore, all require a tight coupling to a test and evaluation apparatus. All also rely on knowing operational details such as goals, mission parameters, and behavioral triggers. Yet, such information is not available when encountering potential autonomous systems in the wild. Yet, there are two reasons measuring level of autonomy based on observed behavior is a critical need. On one hand, there is still a need to compare capabilities between two or more autonomous system. Existing measures provide this albeit during design or test and evaluation phases. Which leads to the need on the other hand- assessing to the degree or level of autonomy the autonomous system actually exhibits relative to mission goals and intended capability. Despite this gap in the research, we uncovered a potential clue in the literature. 

Wolschke et al. (2017) determined evaluation of AS- vehicles, specifically- required knowledge of the systems goals and capabilities. The authors found standard test-case creation based on goals and capabilities insufficient due to the increasing levels of autonomy in such vehicles. As a response, Wolschke et al. demonstrated an observation-based approach to generating AS test-cases. 

This approach uses observation of an action sequence as input, computes the difference between action sequence time-steps using Damerau-Levenshtein string edit distance calculation, and then evaluates the action against the output relative to the expected action sequence. While the output is not a measure of level of autonomy, we think the method points to one potential way to use observed behavior as an input. 

\section{Measuring Observable Level of Autonomy}

We propose the following theoretical framework for a measure for observable level of autonomy. The goal of the framework is to predict a normalized level of autonomy as described in the SAE scale. We designed the framework based on two components: a human behavior lookup table and a foundational expression. Both components build upon existing literature, chiefly Wolschke et al. (2017) and Meakin (2021). 

\subsection{Human behavior lookup table}

The measure requires an accessible lookup table of human behavior exists compatible with the observation. For instance, a turn might consist of the turn direction (left or right), the velocity of the turn, the time spent in the turn (duration), and the angle of the turn (Table 2). Each value is independent from others in the table. Value data types are floating points ($r \in R$). Each individual value contributes necessarily to the overall behavior as subactions. Thus, turning left can be described as $T_l=\{T_v,T_d,T_a\}$.

%% insert table 2

\begin{table}
    \centering
    \begin{tabular}{cc}
        \hline
        \textbf{Value} & \textbf{Description} \\
        \hline
        \hline
         $T_l$ & turn left \\
         $T_r$ & turn right \\
         $T_v$ & turn velocity \\
         $T_d$ & turn duration \\
         $T_a$ & turn angle \\
         \hline
    \end{tabular}
    \caption{Lookup table example for turning.}
    \label{tab:tab2}
\end{table}

We imagine such lookup tables to be application specific in early prototyping. Meaning, the action described in Table 2 is tightly coupled to turning. Moreover, the action would be specific to say an automobile turning. The same table would not apply to a UAV. Therefore, complex behaviors would necessarily include a plethora of lookup tables. However, a generalized methodology for constructing lookup tables should be possible, thus reducing complexity in the measure and increasing operational efficiency.

\subsection{Foundational expression}

Whether the lookup table is tightly coupled to an action or generalized, the foundational expression embodies a central idea: any given action at a specific time step can be treated as a discrete subaction. Further, each discrete subaction can be converted into string element. Thus, we can let $O_a$ be the set of observed actions as $\{o1, o2...on\}$ which are ingested through sensors. The set of human behaviors are ingested from a lookup table (e.g., Table 2) and encoded into $H_a$ as $\{h1, h2, ...hn\}$. For clarity, each element in $O_a$ and $H_a$ is a $T_l$ action equivalent, not a subaction such as $T_v$. 

We then describe the observed level of autonomy using the following expression (Expression 1).

\begin{equation}
    L \equiv \epsilon ( \; (\:\sum_{t=0}^{T} H_a),\; (\:\sum_{t=0}^{T} O_a) \;)
\end{equation}

The level of autonomy $L$ is equivalent to the edit distance function $\epsilon$.  We envision the edit distance function instantiated as equation 6 in Lopresti and Zhou (1996). Total time is $T$ and $t$ is a given time step where $t \in T$ which begins with a zeroed origin representing the initiation of observation. To differentiate, the $T$ time is external whereas any similar symbol part of the human behavior lookup table is internal only. 

\subsection{Algorithmic implementation}

We can describe the foundational expression in conceptual terms using an algorithmic implementation. Meaning, the algorithmic implementation is intended to represent a sequential logic, in a loose narrative form, leading towards Expression 1. This implementation is not representative of how measuring level of autonomy based on observable system behavior ought to be practically implemented. 

In simple terms, the algorithm would be triggered at the start of observing. Then, each discrete observation would be checked against the human behavior lookup table. The observation would be added to the set of observations as long as such is present in the human lookup table. At this step, the subaction value (i.e., $T_v$ or turn velocity) is not considered. Rather, the sets of actions are tested for equivalency after observation has ended. Equivalency is necessary at the action level otherwise the edit distance calculation will fail comparatively. The edit distance is then computed if the set of human actions from the lookup table is the same as the observed actions.

%% insert algorithm
\begin{algorithm}[H]
\SetAlgoLined
\caption{Algorithm 1 Measuring level of autonomy based on observation}
    \KwData{$H_a$}
    \KwResult{Observational Score (0.0 - 5.9)}
    \textbf{WHILE} observing: \\
        \Indp \textbf{FOR} observation $o$ \\
            \Indp lookup $o$ in $H_a$ \\
                \Indp \textbf{IF} true then \textbf{APPEND} $o$ to $O_a$ \\

    \Indm \Indm \Indm \textbf{IF} $H_a$ $\equiv$ $O_a$: \\
    \Indp \textbf{COMPUTE} $\epsilon$
\end{algorithm}

\subsection{Algorithm output and level of autonomy}

The Observational Score output from Algorithm 1 is mapped to a minimum level of autonomy. A minimum is asserted because observation cannot accurately infer a maximum. Moreover, we opt for the five-point SAE level of autonomy scale for illustrative purposes. The Observational Score could be transposed to a range of zero to 10 if desired and mapped to the ALFUS ten-point scale.

%% table 3

\begin{table}
    \centering
    \begin{tabular}{cc}
        \hline
        \textbf{Level of Autonomy} & \textbf{Obervational Score} \\
        \hline
        \hline
         1 & 0.0 - 1.9 \\
         2 & 2.0 - 2.9 \\
         3 & 3.0 - 3.9 \\
         4 & 4.0 - 4.9 \\
         5 & 5.0 - 5.9 \\
         \hline
    \end{tabular}
    \caption{Map of level of autonomy to measure output.}
    \label{tab:tab3l}
\end{table}

\section{Conclusions}

Autonomy research is steadily incorporating AI (Albrecht \& Stone, 2018) as a way to expand operational goals into deeper fields of uncertainty. Such a transition is visible in the shift from AS to AIS (Reis et al., 2021; Chen, Sun, \& Wang, 2022). The motivation driving the AIS shift the need to design systems capable of higher levels of autonomy. This appears to be in response to a selective pressure created greater amounts of uncertainty in operational environments such as driving and national defense.

However, whether an AIS is driving an automobile or selecting targets for a weapon system, keeping a human in the loop or on the loop is required for safety, accountability, and explainability. In this context, the extent of human interaction and interdiction possible translates to a level of autonomy (Hudson \& Reeker, 2007; Meakin, 2021). In simple terms, the higher the level of autonomy the less a human is needed. Principally, there are two level of autonomy scales: a five-point scale (SAE, 2016) and a ten-point scale (Huang et al., 2007). Both scales allow designers to align intended system capabilities with discrete measures. Further, the scales allow for comparison between two or more systems. Such implies a means to measure the level of autonomy.

Therein exists a problem because according to Antsaklis and Rahnama (2018) because measuring a level of autonomy in an AIS is challenging. To this point, while several (Table 1) proposed measures exist, these share a common set of presuppositions and assumptions. Chiefly, the designed goals and capabilities are known to the measures. Further, the measures rely upon access to internal system states. During design and testing, these presuppositions and assumptions are valid. Yet, in the wild, such become demonstrably false. Such inability to measure level of autonomy in the wild is a specific problem because emergent runtime behavior by an AIS may be less than or greater than the design-stated level of autonomy. Neither case is desirable. Accordingly, the purpose of this study was to develop a measure for predicting the level of autonomy in an observed AIS.

To that end, we developed such a measure. The measure attempted to calculate the similarity between observed actions and known human equivalent actions. This calculation is possible because we conceptualized the array of actions in both instances as sets of string data types. Then, we employed a variation an edit distance calculation outlined in Lopresti and Zhou (1996). The adapted calculation suggested the greater edit distance (i.e., less similar), the more an observed system is considered autonomous. Then, we illustrated an implementation of the measure through an algorithm and use case description.

The proposed solution has significance to researchers and practitioners even as a theoretical construct. The grounded value is trifold: (1) measuring level of autonomy in emergent runtime conditions; (2) comparing systems under runtime conditions; and (3) demonstrating a capability based solely on observational inputs. In the context of researchers then, one or more of these grounded values establishes a foundation for development of conceptual constructs. For practitioners, the expression and algorithmic implementation serve as a basis for prototyping applied solutions.

With all this in mind, the proposed solution has several limitations which need to be overcome if a conceptual prototype is to be produced. Foremost, the proposed framework is dependent upon sensor input. Thus, a competent sensor fusion pipeline would be critical in any implementation. Future work is necessary to investigate optimal pipeline features when fusing disparate sensor types and data availability. Further, the measure also assumes sufficient, but not necessarily perfect, observational data. As such, the framework is vulnerable to biased and possibly adversarial input. Future work should explore methods to incorporate responsible AI guardrails to assure input. 

Autonomic computing is an overlooked component in AIS which might provide further insight into evolving a level of autonomy measure based on observation. While some literature (Andresciani \& Cingolani, 2020; Tsamados \& Taddeo, 2023) clearly designates autonomic computing principles as required for AIS, the bulk of research makes little, if any, reference to any self-* concept. We feel autonomic computing should be explored further for potential insights into observational measures of autonomy.

Additionally, cybersecurity controls related to data integrity might be a desirable exploration. Lastly, the proposed framework only applies to observations for which there is a recorded human-based equivalent. Moreover, we assume the instrument (e.g. computer vision) employed to generate the human lookup table is the same instrument used to make in the wild observations for reliability and validity purposes. Perhaps future exploratory research can investigate the extent to which a generalization function may be possible as a replacement for hardcoded human-based lookup tables. Generative AI systems might be a productive line of inquiry in this context.

\section*{Acknowledgments}
I would like to thank my work colleague, Shaho Alaee, for providing insights and feedback during the development of this work.

%Bibliography
\nocite{*}
\bibliographystyle{apalike}
\bibliography{references}

\end{document}